\documentclass[letterpaper, 10 pt, conference]{ieeeconf}  

\IEEEoverridecommandlockouts                              

\usepackage{times}
\usepackage{epsfig}
\usepackage{graphicx}
\usepackage{amsmath}
\usepackage{amssymb}
\usepackage{multirow}
\usepackage[flushleft]{threeparttable}
\usepackage{tikz}
\newcommand{\modelname}{ORStereo}

\newcommand{\reffig}[1]{Fig.~\ref{#1}}
\newcommand{\reftab}[1]{Table~\ref{#1}}
\newcommand{\refeq}[1]{\eqref{#1}}
\newcommand{\tensor}[1]{\mathbf{#1}}
\newcommand{\mss}[1]{\mathrm{#1}}
\newcommand{\globalverticalshrink}{\vspace{-0.1in}}

\usepackage[pagebackref=true,breaklinks=true,colorlinks,bookmarks=false]{hyperref}

\newcommand{\eg}{\emph{e.g}} 
\newcommand{\ie}{\emph{i.e}} 
 
\newcommand{\wrt}{w.r.t}
\newcommand{\etal}{\emph{et al}}

\title{\LARGE \bf
ORStereo: Occlusion-Aware Recurrent Stereo Matching for 4K-Resolution Images}

\author{Yaoyu Hu$^{1}$, Wenshan Wang$^{1}$, Huai Yu$^{1}$, Weikun Zhen$^{1}$, Sebastian Scherer$^{1}$
\thanks{$^{1}$Yaoyu Hu, Wenshan Wang, Huai Yu, Weikun Zhen, and Sebastian Scherer are with the Robotics Institute, Carnegie Mellon University.
        \{{\tt\small yaoyuh, wenshanw, huaiy, weikunz, basti\}@andrew.cmu.edu}}%
}

\begin{document}

\maketitle
\thispagestyle{empty}
\pagestyle{empty}

\begin{abstract}

Stereo reconstruction models trained on small images do not generalize well to high-resolution data. Training a model on high-resolution image size faces difficulties of data availability and is often infeasible due to limited computing resources. In this work, we present the Occlusion-aware Recurrent binocular Stereo matching (\modelname), which deals with these issues by only training on available low disparity range stereo images. \modelname{} generalizes to unseen high-resolution images with large disparity ranges by formulating the task as residual updates and refinements of an initial prediction. \modelname{} is trained on images with disparity ranges limited to 256 pixels, yet it can operate 4K-resolution input with over 1000 disparities using limited GPU memory. We test the model's capability on both synthetic and real-world high-resolution images. Experimental results demonstrate that \modelname{} achieves comparable performance on 4K-resolution images compared to state-of-the-art methods trained on large disparity ranges. Compared to other methods that are only trained on low-resolution images, our method is 70\% more accurate on 4K-resolution images.

\end{abstract}

\section{Introduction}

High-resolution and accurate reconstruction of 3D scenes is critical in many applications. For example, LiDAR scanners with a $\sim$5mm accuracy are typically used to build models for civil engineering analysis. However, this level of accuracy is often not sufficient for detailed inspections and scanners are too expensive and heavy to be carried or flown by drones. Stereo cameras, on the other hand, are compact, and potentially high-resolution source of 3D maps if the data can be effectively processed. 
Most of the recent high-performance stereo matching models are learning-based, however, a small number of them are focusing on high-resolution images.

Recent high-resolution oriented models need dedicated training data to learn the ability to operate large disparity ranges or resort to a combination of models. Yang \etal.\cite{Yang_2019_CVPR_Hierarchical} developed a deep-learning model (HSM) that can handle a disparity range of 768 pixels. They collected a high-resolution (2056$\times$2464) dataset to help the training. HSM has impressive performance on the Middlebury dataset \cite{scharstein2014high} which has a relatively larger image size than other public benchmarks. For higher resolution such as 4K, Hu \etal.\cite{hu2020deep} combined the SGBM~\cite{OpenCVSGBM} method with a deep-learning model to deal with the high-resolution input.

Data availability and limited computing resources are the main issues for training a high-resolution model. Typical binocular stereo datasets have an image size of fewer than 2 million pixels (\eg. Scene Flow~\cite{mayer2016large} $540 \times 960$ or a recent one~\cite{Yang_2019_CVPR_DrivingStereo} with 1762$\times$800) and a limited disparity range (\eg. $\sim$200 pixels). They are far from what we face when using real-world high-resolution images, \eg. 4K-resolution with a disparity range of over 1000 pixels. The next challenge is computing resources. Most deep-learning models consume considerable amounts of GPU memory during training on a small-sized image, \eg. the AANet \cite{Xu_2020_CVPR_AANet} needs 2GB per sample with a crop size of $288 \times 756$ and a disparity range of 192 pixels. However, a higher resolution such as 4K resolution may require a crop width of over 2000 pixels to effectively cover a disparity range of 1000 pixels. It is hard to train a model with high-resolution data directly on typical GPUs since the memory consumption scales approximately in cubic with image dimension and disparity range.

\begin{figure*}[t]
\begin{center}
\includegraphics[width=0.9\linewidth]{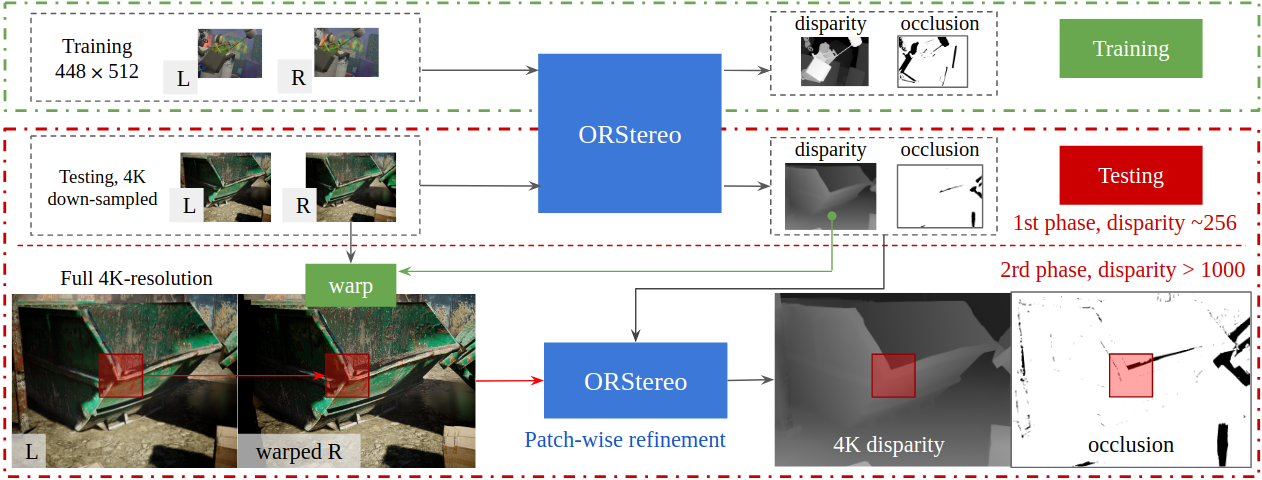}
\end{center}
\globalverticalshrink
\caption{\modelname{}: Occlusion-aware Recurrent binocular Stereo matching. The whole model is trained with low-resolution data. During high-resolution testing, \modelname{} in used in two phases. The down-sampled disparity and occlusion are estimated in the first phase. These predictions are up-sampled and patch-wise refined in the second phase.}
\vspace{-0.1in}
\globalverticalshrink
\label{fig:two_phases}
\end{figure*}

To this end, we choose to not rely on any high-resolution data. Our goal is trying to answer this question: \textbf{how can a model that is trained on low-resolution data be generalized to high-resolution images?} 
Our philosophy is learning how to incrementally refine the disparity instead of predicting it directly. Disparity refinement is less dependent on the size of the input data. Also, it may be possible to trade time with accuracy by performing the refinement on a smaller scale but multiple times. We propose to handle high-resolution data in a two-phase fashion. The first phase results in an initial down-sampled disparity map. In the second phase, the same model recurrently refines the full-resolution disparity in a patch-wise manner. 

While patch-wise processing is widely used for tasks such as object detection, applying a similar strategy for stereo matching faces two issues. First, the small portion of occluded area gets enlarged in some patches, where occluded regions cover most of those patches. These occlusion regions do not have match in stereo images and disturb the refining process. \modelname{} explicitly detects occlusions and stabilizes the recurrent updates. Second, the disparity range in the patch is still as large as it is in the original image, which is on one hand, out of the training distribution, and exceeded the patch size on the other. We find that by utilizing proper normalization techniques, a model can learn the ability to generalize to unseen disparity ranges.


Our model is trained on publicly available datasets with small image sizes. For high-resolution evaluations, we collected a set of 4K-resolution stereo images from both photo-realistic simulations and real-world cameras. The main contributions are summarized as follows.
\begin{itemize}
    \itemsep0em
    \vspace{-0.03in}
    \item  We propose a two-phase strategy for high-resolution stereo matching using regional disparity refinements.
    \item We design a novel structure to recurrently update the disparity and occlusion predictions. The occlusions are explicitly predicted to guide the updates.
    \item We develop a new local refinement module equipped with special normalization operations.
    \item We collect a set of 4K-resolution stereo images for evaluation. This dataset is publicly available from the project web page\footnote{https://theairlab.org/orstereo}.
\end{itemize}

\section{Related work}

Stereo matching is a widely studied topic and there are many works both from geometry-based~\cite{taniai2017continuous} and learning-based research~\cite{Kendall_2017_ICCV, chang2018pyramid, Yao_2018_ECCV}. Many 3D perception tasks have a similar nature with binocular stereo reconstruction, such as multi-view stereo, monocular depth estimation, and optical flow. We find many valuable inspirations and insights from all those tasks.


\textbf{Residual prediction and recurrence}
As stated previously, \modelname{} learns to refine the disparity in a patch multiple times to obtain better accuracy. This process is similar to the residual prediction and recurrence. Many existing works train models to predict residuals to refine an initial prediction~\cite{Pang_2017_ICCV,song2018edgestereo,khamis2018stereonet,Chen_2019_ICCV_Point,dovesi2020real}. However, these models update an initial prediction by a fixed number of steps or dimension scales. In contrast, we introduce disparity improvements with variable steps in a recurrent way.
Most of the recurrent models for 3D perception are designed for dealing with sequential data, \eg. multi-view stereo models\cite{Yao_2019_CVPR_Recurrent, Liu_2020_CVPR_Novel}. As for binocular stereo, Jie \etal.\cite{jie2018left} utilized a recurrent neural network (RNN) to keep track of the consistency between the left and right predictions. 
Recently, RAFT applies an RNN to update optical flow prediction in a way similar to an optimizer, thus the update converges as it proceeds~\cite{teed2020raft}.
Inspired by RAFT, we design a model to recurrently update the disparity prediction without deterioration. To handle the occlusion issue in a small patch from a large image, our model recurrently handles the occlusion.

\begin{figure*}[t]
\begin{center}
\includegraphics[width=0.9\linewidth]{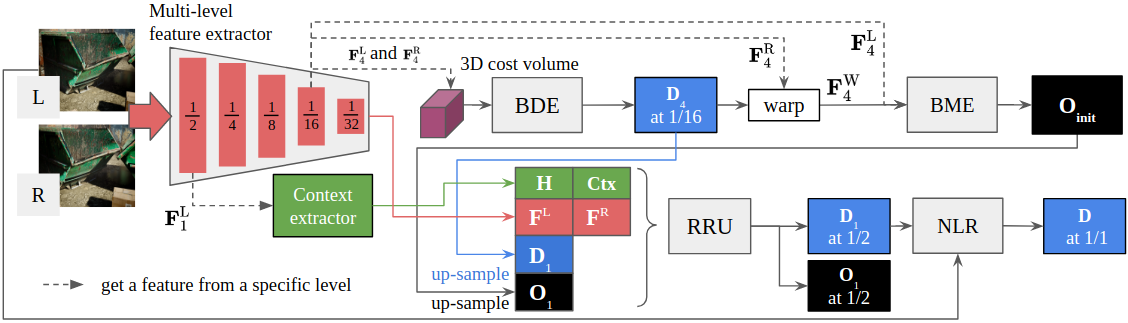}
\end{center}
\globalverticalshrink
\caption{Structure of \modelname. We are using the same structure in both the first and second phases. $\tensor{F}$: feature tensor. $\tensor{D}$: disparity tensor. $\tensor{O}$: occlusion mask. $\tensor{Ctx}$: context information. $\tensor{H}$: hidden state for the recurrent update. Superscripts: L (left), R (right), and W (warped). Subscripts: the level numbers. BDE: base disparity estimator. BME: base occlusion mask estimator. RRU: recurrent residual updater. NLR: normalized local refinement. }
\globalverticalshrink
\label{fig:first_phase}
\end{figure*}


\textbf{Occlusion prediction}
Properly handling occlusion is important for unsupervised learning models, \eg.\cite{Wang_2018_CVPR_occlusion, li2018occlusion}. For supervised models, Ilg \etal.\cite{Ilg_2018_ECCV} demonstrated that a CNN (convolutional neural network) can jointly estimate disparity and occlusion. 
We train \modelname{} in a supervised manner. To deal with large occluded areas in small patches, we chose to explicitly identify occlusions and let the model exploit this information. A recent work \cite{Zhao_2020_CVPR} shares a similar idea where occlusions are used to filter the extracted features before constructing the cost volume. 




\section{Method}
\vspace{-0.02in}


The \modelname{} is trained on low-resolution images and tested on 4K images (\reffig{fig:two_phases}). It consists of 5 components (\reffig{fig:first_phase}): a multi-level feature extractor (FE), a base disparity estimator (BDE), a base occlusion mask estimator (BME), a recurrent residual updater (RRU), and a normalized local refinement component (NLR). When training on small images (\reffig{fig:two_phases} Training, \reffig{fig:first_phase}), BDE and BME learn to predict an initial disparity image and an occlusion mask, which are feed into the RRU as initial values. The RRU works in a recurrent fashion gradually promotes prediction accuracy. At last, the NLR further refines the RRU's prediction and outputs the final results. When testing on 4K image, we use the learned model in a two-phase manner. In the first phase, our model estimates down-sampled disparity and occlusion mask. Then in the second phase (\reffig{fig:two_phases} Testing), the RRU and NLR are reused to refine the enlarged patches in the original resolution. The key features of \modelname{} is that the RRU and NLR are designed to be generalizable to large disparity ranges. As a result, in testing time, when the 4K images with a large disparity range are presented, the RRU and NLR will be able to improve the estimation accuracy even the disparity range is not seen in the training time. 

\subsection{Initial disparity and occlusion estimations}
We place two dedicated components, namely the base disparity estimator (BDE) and the base occlusion mask estimator (BME), to calculate initial values for the RRU. The left and right images are first passed through a five-level feature extractor, which has a simplified ResNet~\cite{He_2016_CVPR_resnet} structure. The extracted features are denoted as $\tensor{F}$ in \reffig{fig:first_phase}. The BDE and BME take the $\tensor{F}_{4}$ (Level 4) features at 1/16 of the input size. The BDE is implemented based on the 3D cost volume technique similar to \cite{Yang_2019_CVPR_Hierarchical}. 
The BME predicts a initial occlusion mask for the RRU. It is designed as a encoder-decoder structure with skip connections. As shown in \reffig{fig:first_phase}, let $\tensor{D}_{i}$ be the disparity tensor at Level $i$, we warp $\tensor{F}^{\mss{R}}_{4}$ by $\tensor{D}_{4}$ to get $\tensor{F}^{\mss{W}}_{4}$. By comparing the feature pattern between $\tensor{F}^{\mss{L}}_{4}$ and $\tensor{F}^{\mss{W}}_{4}$, the BME predicts the initial occlusion mask $\tensor{O}_{\mss{init}}$. 
In testing time, BDE and BME is only used in the first phase. In the second-phase, the initial values for the RRU come from the results of the first phase. 

\begin{figure}[htb]
\begin{center}
\includegraphics[width=0.95\linewidth]{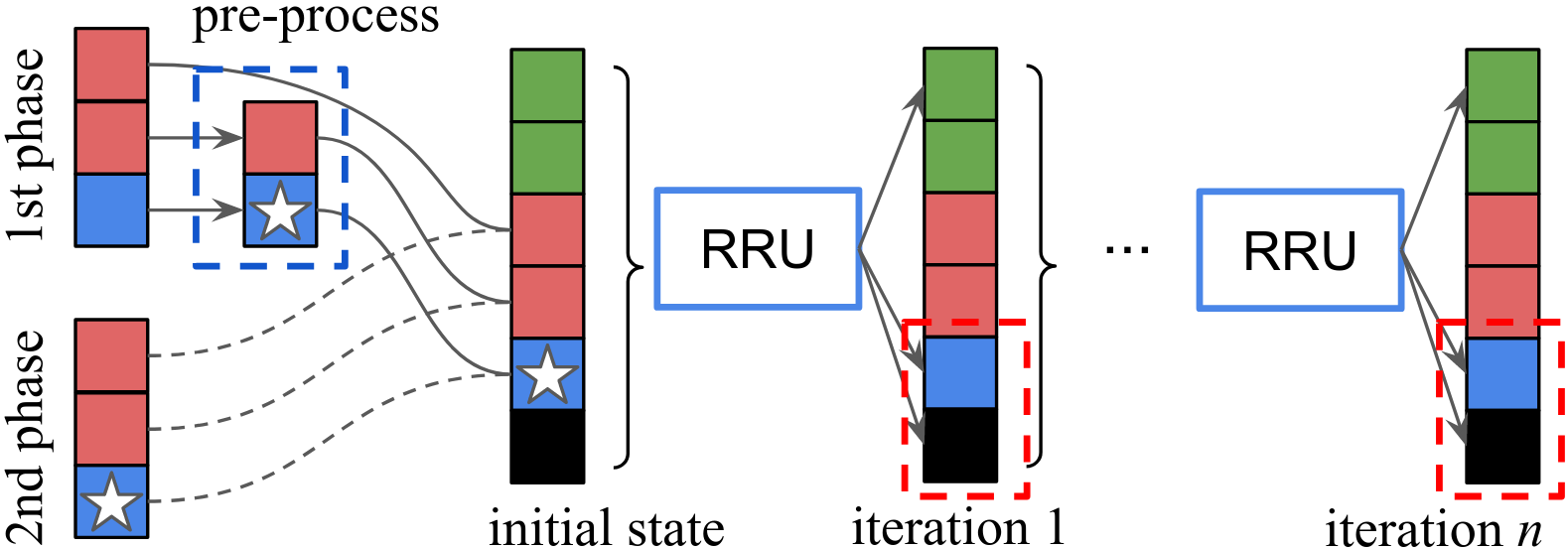}
\end{center}
\globalverticalshrink
\caption{The RRU pre-processing and iteration. Initial state (from top to bottom): the hidden state $\tensor{H}$, context $\tensor{Ctx}$, multi-level features $\tensor{F}^{\mss{L}}$ and $\tensor{F}^{\mss{R}}$, disparity $\tensor{D}_{1}$, and occlusion $\tensor{O}_{1}$.  $\tensor{O}_{1}$ is obtained by resizing the $\tensor{O}$. Blue dashed box: pre-processing in the first phase. Star: zero disparity. Red dashed box: evaluation of the disparity and occlusion losses. The RRU recurrently updates $\tensor{H}$, $\tensor{D}_{1}$, and $\tensor{O}_{1}$. The iteration stops when a fixed $n$ is reached or a convergence criterion triggers.}
\globalverticalshrink
\label{fig:rru_iteration}
\end{figure}

\begin{figure*}[t]
\begin{center}
\includegraphics[width=0.85\linewidth]{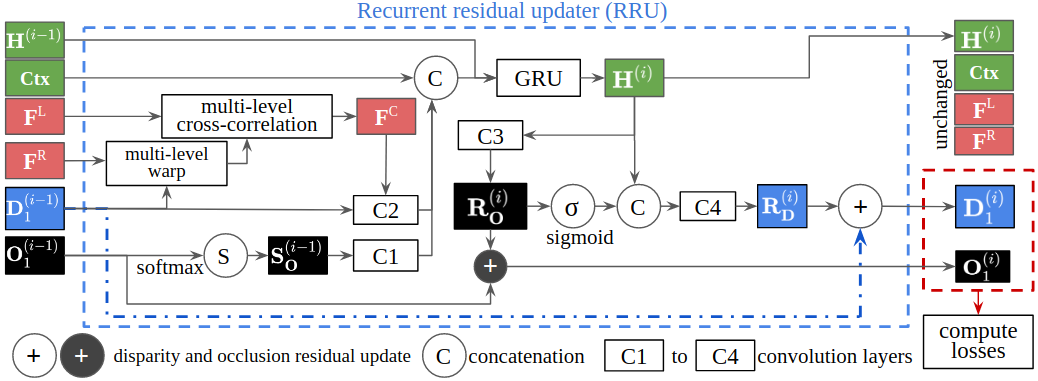}
\end{center}
\vspace{-0.05in}
\caption{The structure of the RRU. Residual update for $\tensor{D}_{1}$ is element wise addition, \refeq{eq:update_disparity}. Residual update for $\tensor{O}_{1}$ is defined in \refeq{eq:update_occlusion}. The multi-level cross-correlation is implemented by referencing \cite{Sun_2018_CVPR}.}
\globalverticalshrink
\label{fig:rru_detail}
\end{figure*}


\subsection{Recurrent residual updater} \label{sec:rru}

To make the RRU work across two phases, we let the RRU do a pre-process on the disparity and always begin from zero disparity. Also, the RRU needs to properly handle occlusions to stabilize the iteration. Note that the RRU takes all the feature levels ($\tensor{F}^{\mss{L}}$ and $\tensor{F}^{\mss{R}}$) and updates disparity and occlusion at Level 1 ($\tensor{D}_{1}$ and $\tensor{O}_{1}$) with 1/2 image width.

\textbf{Initial state with zero disparity.} The RRU uses multiple input types to start the iteration in both phases, as shown in \reffig{fig:rru_iteration}. In the training time, \modelname{} only goes through the first phase. In the second phase as shown in \reffig{fig:two_phases}, the full-resolution right image gets warped before going into \modelname{}. Ideally, the warped image should roughly be the same compared with the left image in non-occluded regions.
Then patches from the left and the warped images should now correspond to disparity values close to zero at non-occluded pixels. Considering this phenomenon, we add a pre-processing in the first phase to warp $\tensor{F}^{\mss{R}}$ and assign an all-zero $\tensor{D}_{1}$ to the RRU (\reffig{fig:rru_iteration}). The RRU always has a zero $\tensor{D}_{1}$ in the second phase.


\textbf{The recurrent iterations.} We design the RRU by augmenting a GRU (gated recurrent unit) similar to \cite{teed2020raft}. The detailed model structure is illustrated in \reffig{fig:rru_detail}. The superscript $(i)$ denotes the iteration step.
For the $i$th iteration, the RRU computes $\tensor{R}^{(i)}_{\tensor{O}}$ and $\tensor{R}^{(i)}_{\tensor{D}}$ as the residuals for occlusion $\tensor{O}^{(i)}_{1}$ and disparity $\tensor{D}^{(i)}_{1}$ based on the GRU's hidden variable $\tensor{H}^{(i)}$. Similar to the \cite{teed2020raft}, we keep a context feature, $\tensor{Ctx}$, as a constant reference to the initial state. The key differences from \cite{teed2020raft} are that \modelname{} recurrently exploits the occlusion information to stabilize the recurrent iteration. 


Since the RRU works in a recurrent way, during an evaluation after training, we can apply it like an optimizer by keeping it updating until a pre-defined criterion is satisfied. Here, unsupervised loss functions for similar perception tasks are reasonable criterion candidates. Later in the experiment section, we tested the SSIM\cite{Godard_2017_CVPR}, which is widely accepted as a robust similarity measure for unsupervised methods. The RRU can keep updating a patch until the SSIM value between the original and warped images goes down. However, we found that the SSIM is not reliable and the RRU simply performs better with fixed and longer iterations.



\textbf{Residual update of occlusions.} Iterations become unstable in occluded regions where little information is useful for stereo matching. To improve robustness, the RRU explicitly handle the occlusion by the occlusion-augmented $\tensor{H}^{(i)}$(shown in \reffig{fig:rru_detail}). 
$\tensor{O}^{(i)}_{1}$ is represented as a 2-channel tensor in \modelname{} and supervised by cross-entropy loss, which does not require a value-bounded $\tensor{O}^{(i)}_{1}$. However, this unbounded value cause problems for long recurrent updates. \reffig{fig:rru_detail} shows that $\tensor{O}^{(i-1)}_{1}$ first passes through a soft-max operator. This ensures an intermediate value-bounded representation, $\tensor{S}^{(i)}_{\tensor{O}}$, and keeps the gradients of the loss value \wrt{} $\tensor{O}^{(i)}_{1}$ from diminishing. Similar reasoning also explains the purpose of the Sigmoid operator, which bounds the value of $\tensor{R}^{(i)}_{\tensor{O}}$. Here, $\tensor{R}^{(i)}_{\tensor{O}}$ is a single-channel tensor. $\tensor{O}^{(i-1)}_{1}$ gets updated by \refeq{eq:update_occlusion}.
\begin{equation} \label{eq:update_occlusion}
    V\left( \tensor{O}^{(i)}_{1}, j \right ) = V\left( \tensor{O}^{(i-1)}_{1}, j \right ) - (-1)^{j} \tensor{R}^{(i)}_{\tensor{O}}
\end{equation}
\noindent where function $V(\cdot,j)$ takes out the channel at index $j$ from a tensor and $j\in\{0, 1\}$. Equation \refeq{eq:update_occlusion} also makes the iteration more stable. Equation \refeq{eq:update_disparity} gives the disparity update.
\begin{equation} \label{eq:update_disparity}
    \tensor{D}^{(i)}_{1} = \tensor{D}^{(i-1)}_{1} + \tensor{R}^{(i)}_{\tensor{D}}
\end{equation}

\subsection{Normalized local refinement}

\begin{figure}[h]
\begin{center}
\includegraphics[width=0.95\linewidth]{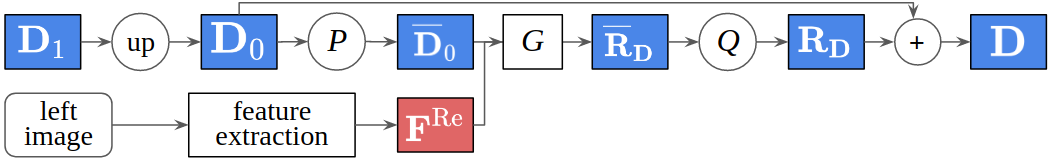}
\end{center}
\globalverticalshrink
\caption{The normalized local refinement module (NLR). Refer to \refeq{eq:P_D0} to \eqref{eq:Q_RD} for definitions of some symbols. }
\label{fig:final_refine}
\end{figure}

Following the RRU, the normalized local refinement module (NLR) adds a final update to the disparity $\tensor{D}_{0}$, as shown in \reffig{fig:final_refine}. The NLR is designed to smooth the object interior and sharpen boundaries by exploiting local consistency between the disparity and the input image.
The NLR extracts its own features from the left image and works directly at the image resolution.

\begin{table*}[bt]
\scriptsize
\caption{EPE metrics of \modelname{} compared with the SotA models on the Scene Flow dataset. 
}
\begin{tabular}{cccccccccc}
\hline
MCUA\cite{Nie_2019_CVPR_Multi} & Bi3D\cite{Badki_2020_CVPR_Bi3D} & GwcNet\cite{Guo_2019_CVPR_Group} & FADNet\cite{wang2020fadnet} & GA-Net\cite{Zhang_2019_CVPR_GA_Net} & WaveletStereo\cite{Yang_2020_CVPR_Wavelet} & DeepPruner\cite{Duggal_2019_ICCV_DeepPruner} & SSPCV-Net\cite{Wu_2019_ICCV_Semantic}    & AANet\cite{Xu_2020_CVPR_AANet}    & \modelname{} (ours) \\ \hline
0.56$^*$ & 0.73 & 0.77$^\blacktriangle$  & 0.83   & 0.84   & 0.84          & 0.86       & 0.87 & 0.87$^*$ & 0.74      \\ \hline
\end{tabular}
\\ \modelname{} only goes through the first phase for this low-resolution test. $*$ Best values from individual works. $\blacktriangle$ Finalpass. \modelname{} is trained and tested on the cleanpass subset of the Scene Flow dataset. Some samples are removed as suggested by the Scene Flow dataset.
\globalverticalshrink
\label{tab:scene_flow}
\end{table*}

The NLR also works in both phases. In the second phase the NLR faces a major challenge as disparity values go beyond the training range. The way the NLR handles this challenge is always working under a normalized disparity range. In \reffig{fig:final_refine}, the up-sampled disparity, $\tensor{D}_{0}$, is normalized as $\overline{\tensor{D}}_{0}$ by function $P$. Then module $G$ processes the feature $\tensor{F}^{\mss{Re}}$ and $\overline{\tensor{D}}_{0}$ to produce residual $\overline{\tensor{R}}_{\tensor{D}}$. Then a de-normalization function $Q$ re-scales $\overline{\tensor{R}}_{\tensor{D}}$ to $\tensor{R}_{\tensor{D}}$. The final disparity $\tensor{D}$ is then obtained in the same way of \refeq{eq:update_disparity}. Module $G$ has an encoder-decoder structure. $P$ and $Q$ are defined from \refeq{eq:P_D0} to \eqref{eq:Q_RD}.
\begin{align}
P( \tensor{D}_{0} ) &= ( \tensor{D}_{0} - m ) / ( s + \epsilon ) = \overline{\tensor{D}}_{0} \label{eq:P_D0} \\
Q\left( \overline{\tensor{R}}_{\tensor{D}} \right) &=  (s + \epsilon) \overline{\tensor{R}}_{\tensor{D}} = \tensor{R}_{\tensor{D}} \label{eq:Q_RD}
\end{align}
\noindent where scalar $m = \mathrm{mean}( \tensor{D}_{0} )$ and $s = \mathrm{std}(\tensor{D}_{0})$ are the mean and standard deviation of $\tensor{D}_{0}$. $P$ normalizes the disparity, and $Q$ de-normalizes the residual value. Here, $W$ serves as a local weight which encourages large residuals near local discontinuities. $\epsilon$ is a constant hyperparameter. $P$ and $Q$ enable the NLR to generalize to large disparity ranges in the second phase by making the NLR work in a normalized disparity range.

\subsection{Training losses}

We adopt a supervised scheme for all the outputs. The total training loss is defined as \refeq{eq:loss}.
\begin{align} \label{eq:loss}
    l^{\mss{total}} & = \lambda^{\tensor{D}}_{4} \mathrm{SL}\left( \tensor{D}_{4}, \tensor{D}^{\mss{t}}_{4} \right) + \lambda^{\tensor{O}} \mathrm{CE}\left( \tensor{O}, \tensor{O}^{\mss{t}} \right) \nonumber \\
    & + \lambda^{\tensor{D}}_{1} \sum^{n}_{i=1} (\gamma^{\tensor{D}})^{n-i+1} \mathrm{SL} \left( \tensor{D}^{(i)}_{1}, \tensor{D}^{\mss{t}}_{1} \right) \nonumber \\
    & + \lambda^{\tensor{O}}_{1} \sum^{n}_{i=1} (\gamma^{\tensor{O}})^{n-i+1} \mathrm{CE} \left( \tensor{O}^{(i)}_{1}, \tensor{O}^{\mss{t}}_{1} \right) \nonumber \\
    & + \lambda^{\tensor{D}} \mathrm{SL}\left( \tensor{D}, \tensor{D}^{\mss{t}} \right)
\end{align}
\noindent where $\mathrm{SL}$ and $\mathrm{CE}$ are the smooth L1 and cross-entropy loss functions. $n$ is the iteration number and from $\lambda^{\tensor{D}}_{4}$ to $\lambda^{\tensor{D}}$ are the constant weights for different loss values (see \reftab{tab:constants}).

\subsection{Working in the second phase}

We make patches out of four objects for the second phase: the left image, the warped right image, the disparity, and the occlusion mask, all in the full-resolution. Our experiments show that keeping small overlaps among patches achieves better results since accuracy may drop near the patch borders. In the overlap region, the disparity and occlusion predictions are averaged across patches.

\section{Experiments}

\subsection{Datasets and details of training}
Our target is 4K-resolution stereo reconstruction with over 1000 pixels of disparity. \modelname{} allows us to train with only small-sized images and a typical disparity range around 200 pixels. We utilize several public datasets for the training, \ie., the Middlebury dataset at 1/4 resolution~\cite{scharstein2014high}, the Scene Flow~\cite{mayer2016large} dataset ($\sim$35k stereo pairs), and the TartanAir~\cite{wang2020tartanair} datasets ($\sim$18k pairs sampled). 
The Scene Flow and TartanAir datasets do not provide true occlusion labels. We generate them by comparing the left and right true disparities. We will not use the KITTI dataset despite its popularity because it is hard to generate reliable dense occlusion labels from sparse true disparities. 

\begin{figure}[ht]
\begin{center}
\includegraphics[width=0.95\linewidth]{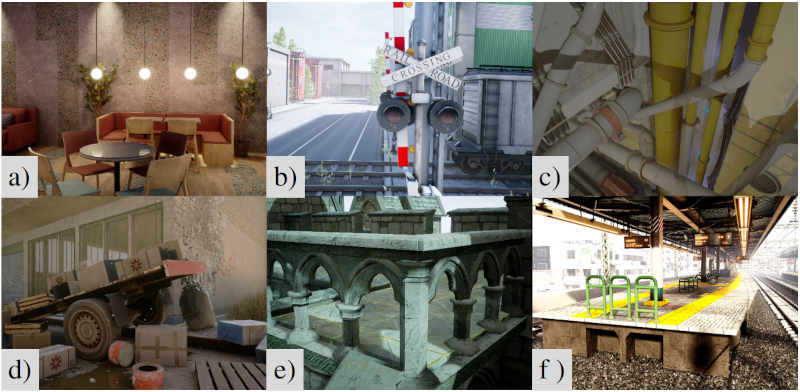}
\end{center}
\globalverticalshrink
\caption{Sample images from the 100 pairs of 4K-resolution stereo images. 6 of 7 environments are shown: a) restaurant, b) factory district, c) under ground work zone, d) city ruins, e) ancient buildings, f) train station.}
\globalverticalshrink
\label{fig:6_samples}
\end{figure}

Currently, no public stereo benchmark provides 4K-resolution data. So we collect a set of synthetic 4K-resolution photo-realistic stereo images with ground truth disparity. These images are captured by AirSim \cite{airsim2017fsr} in the Unreal Engine. We prepare 100 pairs of stereo images from 7 simulated environments. Some cases may have a disparity range of over 1200 pixels. The evaluation cases in \reffig{fig:results_on_our_4k} and \ref{fig:4k_sample} are from this dataset. Additional samples are shown in \ref{fig:6_samples}. The dataset is only used for evaluation and all images and ground truth disparities are available at the project page.

Training only happens in the first phase of \modelname{}. We use a disparity range of 256 pixels and randomly crop the images to $448 \times 512$ pixels. We set the iteration number to 4 for training the RRU. Similar works such as \cite{teed2020raft, Bai_2020_CVPR_Deep} also use fixed number of iterations during training. Other model constants are listed in \reftab{tab:constants}, where the loss weight constants are chosen to balance the values from different loss components. We train \modelname{} with 4 NVIDIA V100 GPUs with a mini-batch of 24 for all experiments. Other training settings varying among different experiments will be shown separately. 

\begin{table}[!ht]
\begin{center}
\caption{The model constants of \modelname{}.}
\label{tab:constants}
\begin{tabular}{ccccccccc}
\hline
$\epsilon$ & $\lambda^{\tensor{D}}_{4}$ & $\lambda^{\tensor{O}}$ & $\lambda^{\tensor{D}}_{1}$ & $\lambda^{\tensor{O}}_{1}$ & $\lambda^{\tensor{D}}$ & $\gamma^{\tensor{D}}$ & $\gamma^{\tensor{O}}$ \\ \hline
1e-6       & 32                         & 2                      & 2                          & 1                          & 2                      & 0.8                   & 0.8                   \\ \hline
\end{tabular}
\end{center}
The values from $\lambda^{\tensor{D}}_{4}$ to $\lambda^{\tensor{D}}$ are selected to make the loss components have similar quantities in the end of a training.
\globalverticalshrink
\end{table}

\subsection{Evaluation the first phase on small images}
To evaluate the first phase of \modelname{} on small-sized images, we train \modelname{} on the Scene Flow dataset only and then compare it with the state-of-the-art (SotA) models. 
We limit the metric computation under 192 pixels, matching the SotA models. \reftab{tab:scene_flow} lists the results on the testing set of the Scene Flow dataset measured in the average EPE (end point error) metric. \modelname{} achieves comparable performance with the SotA models. 

\subsection{Evaluation on high-resolution images}
For better performance and generalization ability across image sizes and disparity ranges, the Middlebury (at 1/4 resolution) and TartanAir datasets are added to the training. We further augment the data with random color, random flip, and random scale similar to \cite{Yang_2019_CVPR_Hierarchical}. After training, we apply a 512$\times$512 patch size with an overlap of 32 pixels in the second phase. In Sec.~\ref{sec:ablations}, after comparing the performance with various iteration numbers, we make the RRU iterate for 10 steps in both the phases for the subsequent evaluations.

Although the training data have small size, we expect \modelname{} to learn the ability to refine a patch of disparity at a higher resolution. We first show results on the Middlebury dataset~\cite{scharstein2014high} with full resolution. This dataset is still considered as hard for models trained on low-resolution data.
We choose three recent models (with available pre-trained weights) that are trained with low-resolution data and have openly evaluated their performance on the Middlebury dataset. Additionally, we include the HSM model\cite{Yang_2019_CVPR_Hierarchical}, which can cover 768 pixels of disparity after training on high-resolution images. 
The quantitative results are listed in \reftab{tab:middlebury}. Trained on smaller image size, we observed that \modelname{} delivers accuracy close to the SotA \cite{Yang_2019_CVPR_Hierarchical} trained on high-resolution images. Later, when the resolution goes up to 4K, \modelname{} can still maintain its performance.

\begin{table}[htb]
\begin{center}
\caption{Comparison on the Middlebury evaluation dataset}
\label{tab:middlebury}
\begin{tabular}{c|cccc|c}
\hline
\begin{tabular}[c]{@{}c@{}}Model\\ \& scale\end{tabular} & \begin{tabular}[c]{@{}c@{}}AANet\\ \cite{Xu_2020_CVPR_AANet} 1/2\end{tabular} & \begin{tabular}[c]{@{}c@{}}DeepPruner\\ \cite{Duggal_2019_ICCV_DeepPruner} 1/4\end{tabular} & \begin{tabular}[c]{@{}c@{}}SGBMP\\ \cite{hu2020deep} full\end{tabular} & \begin{tabular}[c]{@{}c@{}}ORStereo\\ (ours) full\end{tabular} & \begin{tabular}[c]{@{}c@{}}HSM$^{\blacktriangle}$\\ \cite{Yang_2019_CVPR_Hierarchical} full\end{tabular} \\ \hline
EPE                                                     & 6.37                                                 & 4.80                                                        & 7.58                                                 &                3.23                                                & 2.07                                               \\ \hline
\end{tabular}
\end{center}
$\blacktriangle$This model is trained on high-resolution data. \modelname{} achieves near SotA performance without training on high-resolution data. Non-occluded EPE is reported. All EPE values are associated with specific image scales. All values are from the evaluation set and published on the website of Middlebury dataset under the "test dense avgerr nonocc" category.
\end{table}

\begin{figure*}[!htb]
\begin{center}
\includegraphics[width=0.97\linewidth]{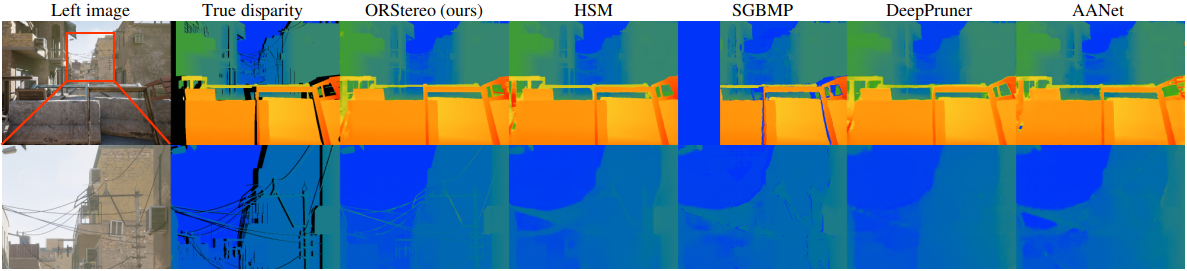}
\end{center}
\globalverticalshrink
\caption{Results on our 4K synthetic stereo images. The color map is the same as \reffig{fig:4k_sample}. Columns 3-7 are the disparity estimation of each model. \reftab{tab:airsim_4k_eval} shows the quantitative statistics over all the 100 evaluation samples. Seeing thin structures is one of the motivation for high-resolution stereo reconstruction. Our model exploits and reconstructs fine-grain details as shown in the zoom-in figures. }
\globalverticalshrink
\label{fig:results_on_our_4k}
\end{figure*}

\begin{figure}[ht]
\begin{center}
\includegraphics[width=0.9\linewidth]{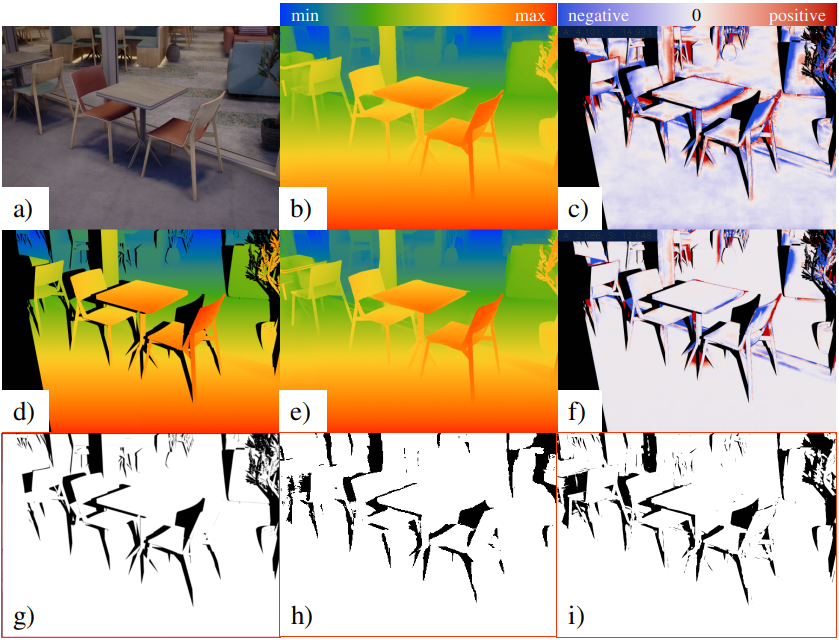}
\end{center}
\globalverticalshrink
\caption{A 4K sample and \modelname{} results. a) 4K Left image. b, c) disparity and error of the first phase. d) ground truth disparity with occlusion. e, f) disparity and error of the second phase. g) true occlusion mask. h, i) occlusion prediction of the first and second phases. c) and f) are masked by the true occlusion from g). The EPE values are 4.10 in c) and 2.09 in f), meaning that the disparity map gets improved in the second phase. }
\globalverticalshrink
\label{fig:4k_sample}
\end{figure}

Some of the models in \reftab{tab:middlebury} are unable to operate 4K-resolution images with a limited memory budget or their effective disparity ranges are not enough, we down-sample the input until a model can handle it. Then the results are re-scaled to the original resolution before evaluation. Besides the EPE metric. Evaluation is done with all the collected samples. \reffig{fig:4k_sample} shows a sample output from \modelname{}. In this figure, we can first compare the disparity predictions from the two phases. The EPE of the first phase is 4.10 pixels (\reffig{fig:4k_sample} c, up-sampled back to 4K-resolution). \modelname{} improves this value to 2.09 pixels after the second phase (\reffig{fig:4k_sample} f). As shown in \reftab{tab:airsim_4k_eval}, although \modelname{} is trained on small image size, its EPE is even lower than HSM on these 4K images with a smaller memory consumption. \reffig{fig:results_on_our_4k} further provides us with a qualitative comparison. \modelname{} has the potential to fully utilize the fine-grained details in high-resolution images and reconstructs thin structures that are better captured by high-resolution images. \modelname{} achieves this at a cost of execution time because it needs to refine multiple patches and the RRU iterates several steps for each patch. On average, \modelname{} spends about 15s for a single 4K-resolution image.

\begin{table}[ht]
\begin{center}
\caption{Comparison on the synthetic 4K dataset. }
\label{tab:airsim_4k_eval}
\begin{tabular}{lllll}
\hline
Model                                                       & Scale & Range & EPE                    & Mem (MB)      \\ \hline
AANet\cite{Xu_2020_CVPR_AANet}                              & 1/8   & 192   & 9.96                  & 8366      \\
DeepPruner\cite{Duggal_2019_ICCV_DeepPruner}                & 1/8   & 192   & 8.31                   & 4196      \\
SGBMP$^*$\cite{hu2020deep}                                  & 1     & 256   & 4.21                  & 3386          \\
\modelname{} (ours)                                                   & 1     & 256   & \textbf{2.37} & \textbf{2059} \\ \hline
HSM$^{\blacktriangle}$\cite{Yang_2019_CVPR_Hierarchical} & 1/2   & 768   & 2.41              & 3405          \\ \hline
\end{tabular}
\end{center}
\modelname{} shows the best results among the models trained on small images. There are 100 pairs of stereo images. Scale: the down-sample scale against the original image width. Range: the trained disparity range. EPE: non-occluded EPE. Mem: peak GPU memory (first/subsequent). $*$SGBMP combines a learning-based and a non-learning-based model, only GPU memory consumption is reported. $\blacktriangle$This model is trained on high-resolution data.
\globalverticalshrink
\end{table}

\begin{figure*}[!ht]
\begin{center}
\includegraphics[width=1.0\linewidth]{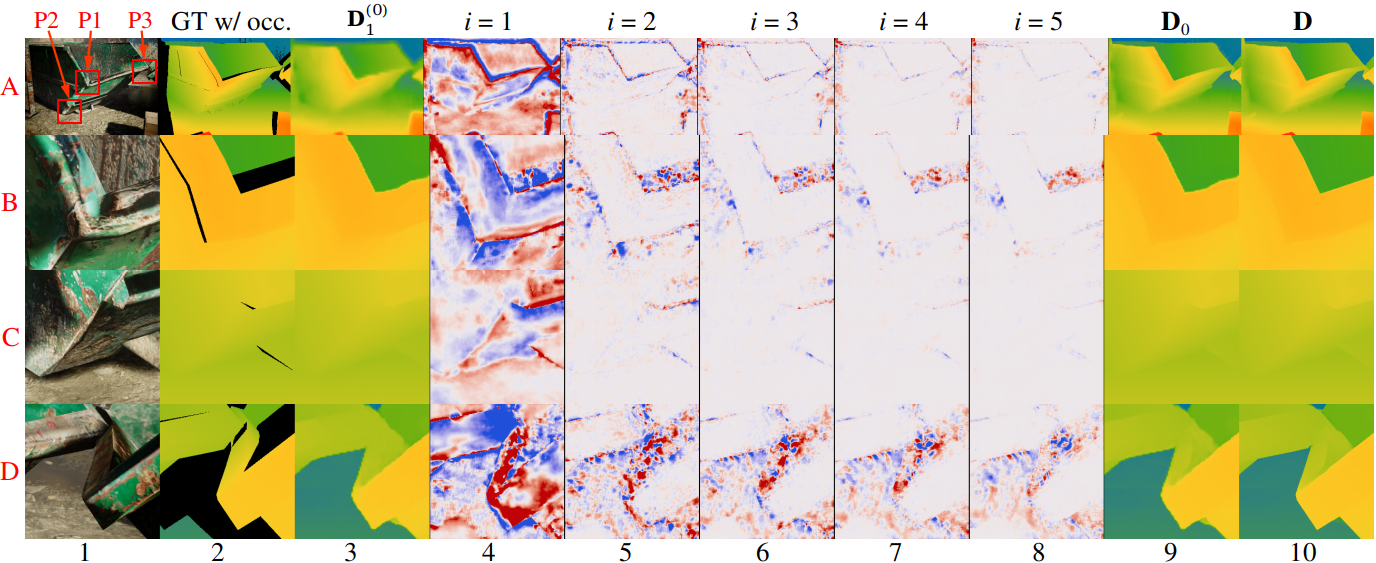}
\end{center}
\globalverticalshrink
\caption{The two phases of \modelname{}. The color maps are the same as \reffig{fig:4k_sample}. Input: 4K-resolution. Row A: the first phase. 3 patches (512$\times$512, P1-P3) are cropped out for illustrating the second phase (Row B-D). Row B (P1): moderate disparity change and occlusion. Row C (P2): short disparity range, limited occlusion. Row D (P3): large disparity jump, severe occlusion. Column 2, 3, 9, 10 are disparities. Column 4-8 show the first 5 updates made by the RRU. The RRU quickly converges in the non-occluded regions after several iterations and oscillates in the occluded areas. The NLR (Column 10) smooths and sharpens $\tensor{D}_{0}$ (Column 9) in both phases. The results also indicate that the NLR can handle unseen disparities beyond the training range.}
\label{fig:iterations}
\end{figure*}

\subsection{Ablation studies} \label{sec:ablations}
For a better understanding on how various factors affect the performance of \modelname{}, we additionally train two models: one that has no explicit occlusion treatments (with all the occlusion update shown in \reffig{fig:rru_detail} removed), and another one with less training data (remove all samples from TartanAir). Furthermore, we test various iteration numbers and disable the RRU or NLR. All the above model variants are evaluated on the same 100 pairs of 4K stereo images used in \reftab{tab:airsim_4k_eval}. The results are shown in \reftab{tab:ablations}. We first observe that simply making the RRU iterate longer than the training setting (4 steps) improves the overall accuracy. Thus the RRU learns to work like an optimizer, which gradually improve an initial prediction. However, this does not hold with extensively long iterations, \eg. model F$_{\mss{20}}$. The F$_{\mss{S}}$ model shows that monitoring the RRU by the SSIM is not reliable. This may be due to the fact that SSIM fails to distinguish stereo matching in texture-less and repeated texture regions. The model variants (V$_{1}$-V$_{4}$) regarding to less training data and incomplete model structures all experience a performance drop, meaning that \modelname{} does benefit from its special design. Notably, the V$_1$ model that is trained with less amount of data than the SotA methods in \reftab{tab:airsim_4k_eval} still achieves the best EPE among the models trained on low-resolution data.

\begin{table}[htb]
    \scriptsize
    \begin{center}
    \caption{Performance comparison based on various factors.}
    \label{tab:ablations}
    \resizebox{0.48\textwidth}{!}{%
\begin{tabular}{c|ccccccccc}
\hline
Model     & F$_4$      & F$_{10}$      & F$_{15}$   & F$_{20}$   & F$_{\mathrm{S}}$ & V$_1$      & V$_2$      & V$_3$      & V$_4$      \\ \hline
All data  & \checkmark & \checkmark    & \checkmark & \checkmark & \checkmark & -          & \checkmark & \checkmark & \checkmark \\
Occlusion & \checkmark & \checkmark    & \checkmark & \checkmark & \checkmark & \checkmark & -          & \checkmark & \checkmark \\
RRU Iter. & 4          & 10            & 15         & 20         & SSIM       & 10         & 10         & -          & 10         \\
NLR       & \checkmark & \checkmark    & \checkmark & \checkmark & \checkmark & \checkmark & \checkmark & \checkmark & -          \\ \hline
EPE       & 2.43       & \textbf{2.37} & 2.45       & 2.56       & 2.55       & 3.34       & 2.70       & 11.46       & 3.34       \\ \hline
\end{tabular}%
}
    \end{center}
    F4-F10: Full model with various RRU iteration numbers. F4 is the training setting. F10 is the model used in \reftab{tab:middlebury} and \ref{tab:airsim_4k_eval}. F$_{\mss{S}}$ utilzes the SSIM to monitor RRU iterations. V1-V4: model variants. V1 is trained without the TartanAir dataset, making the training set have less samples than the other SotA models shown in \reftab{tab:airsim_4k_eval}. 
\end{table}

\subsection{The characteristics of the two-phase procedure} \label{sec:the_two_phases}
The RRU and NLR are the core components for \modelname{} to work across phases. To illustrate the characteristics of the RRU and NLR, in \reffig{fig:iterations}, we show their behaviors in a complete first phase and 3 patches from the second phase. \reffig{fig:iterations} shows that the RRU delivers stable and convergent updates in the non-occluded areas especially Row D in the figure. In occluded regions, there is hardly any useful information for stereo matching. However, inside a severely occluded patch, \modelname{} can use the detected occlusion to guide the residual update, stabilizing the updates in the non-occluded regions. 


\subsection{Evaluation on real-world 4K stereo images}
\begin{figure}[htbp]
    \begin{center}
    \includegraphics[width=1.0\linewidth]{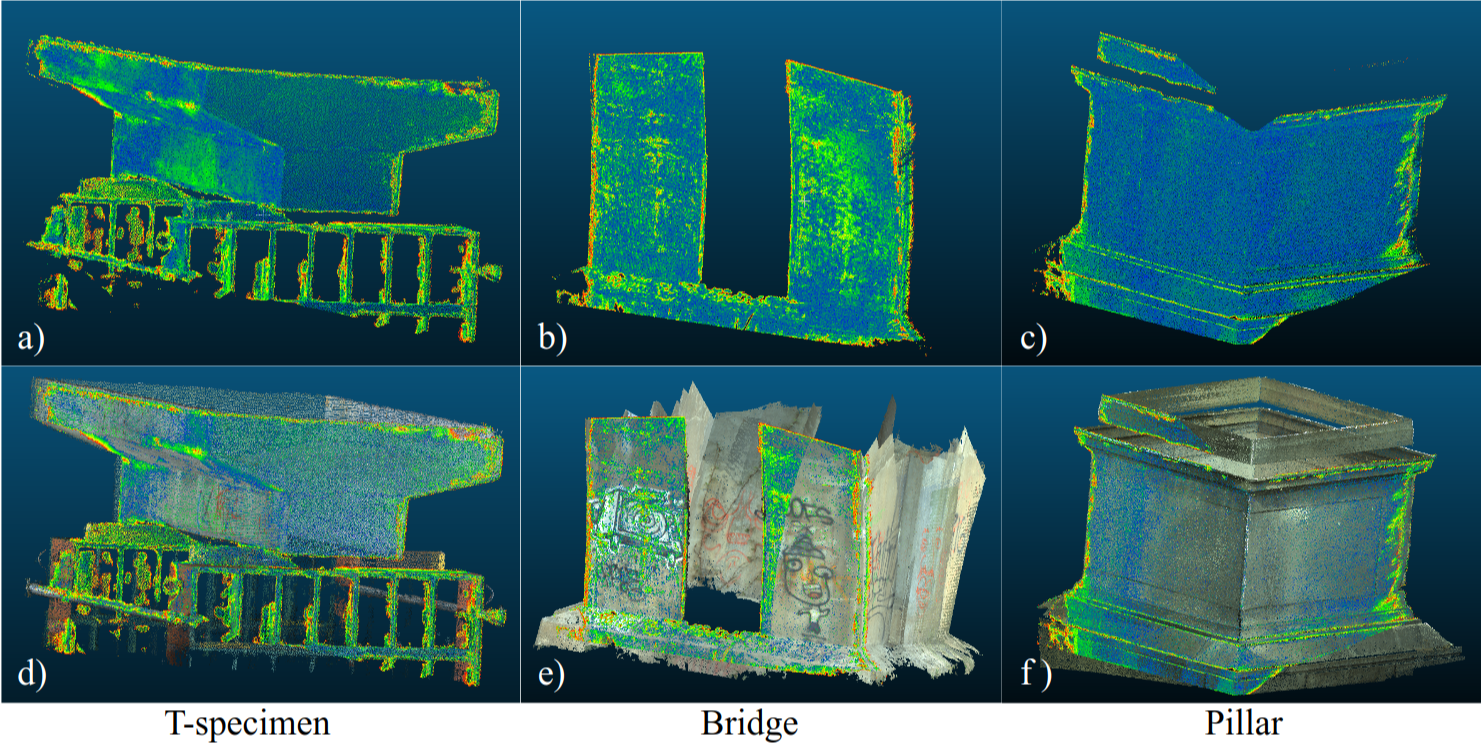}
    \end{center}
	\globalverticalshrink
	\caption{Reconstruction results aligned with ground truth. Overlapping points are colorized by the point-to-plane distance to the ground truth by blue-green-yellow-red in an increasing manner. a)-c): points from \modelname{}. d)-f): the top row with the ground truth.}
	\globalverticalshrink
	\label{fig:gtcompare}
\end{figure}

Evaluation on real-world high-resolution data is necessary because \modelname{} is trained with only synthetic images. We set up a customized stereo camera with two 4K cameras and a LiDAR and capture images around various objects. To evaluate the stereo reconstruction results, we generate dense point clouds as ground truth using a LiDAR-enhanced SfM \cite{zhen2020lidar} method. Using point clouds allows us to measure the metric error of the reconstructed model. For each scene, all disparity maps from different viewpoints are converted to point clouds and are registered with the ground truth by the ICP method with an initial pose guess from the SfM results. Both the ground truth and stereo reconstructed point clouds are down-sampled for efficiency. The results are shown in \reffig{fig:gtcompare} and \reftab{tab:gtcompare} with three different scenes. We use the mean point-to-plane distance ($rmse$) of all the overlapping points between stereo point clouds and ground truth to measure the precision. A point is considered to have a valid overlap if it falls within 0.1m from any ground truth point. Additionally, we use the number of overlapping points $num$ as a metric to measure the valid overlapped area. Both HSM\cite{Yang_2019_CVPR_Hierarchical} and \modelname{} achieve similar millimeter-level precision. 
Notably, \modelname{} obtains higher $num$ than HSM, which means our model reconstructs more valid areas. \modelname{} is trained on small-sized synthetic data and utilizes fewer computation resources but still obtains competitive reconstruction results on real-world 4K stereo images.

\begin{table}[htbp]
    \small
	\begin{center}
	\caption{Stereo reconstruction results on real-world 4K images.}
	\label{tab:gtcompare}
	\setlength{\tabcolsep}{1.0mm}{
	\begin{tabular}{c|cc|cc|cc}
	\hline
	Scene & \multicolumn{2}{|c}{\text{T-specimen}} & \multicolumn{2}{|c}{\text{Pillar}}& \multicolumn{2}{|c}{\text{Bridge}}  \\
	Frames & \multicolumn{2}{|c}{25} & \multicolumn{2}{|c}{29} & \multicolumn{2}{|c}{32} \\
	Metrics& $rmse$ & $num$ & $rmse$ & $num$ & $rmse$ & $num$  \\
	\hline
	HSM\cite{Yang_2019_CVPR_Hierarchical}  & 6.56   & 1.30  & 3.94  & 2.43 & 5.82 & 3.50   \\ 
	\modelname{} & 6.74 & \textbf{1.37} & \textbf{3.69} & \textbf{2.48} & \textbf{5.33} & \textbf{3.62}   \\
	\hline
	\end{tabular}}
	\end{center}
	Frames: the total number of stereo pairs. $rmse$: point-to-plane distance in millimeters. $num$: valid overlapping points, unit $10^5$. $rmse$ and $num$ are averaged across all the frames of individual cases. Evaluations are conducted by down-sampling the point clouds by a grid size of 5mm.
	\vspace{-0.2in}
\end{table}  


\section{Conclusions}
We present \modelname{}, a model that is trained only on small-sized images but can operate high-resolution stereo images at inference time with limited GPU memory. In testing time, \modelname{} refines an initial prediction in a patch-wise manner at full resolution. By jointly predicting the disparity and occlusion, the recurrent residual updater (RRU) can steadily update a disparity patch with severe occlusions. With a special normalization and de-normalization sequence, the normalized local refinement module (NLR) can generalize to unseen large disparity ranges. Our experiments on synthetic and real-world 4K-resolution images validate the effectiveness of \modelname{} in both low- and high-resolution stereo reconstruction. \modelname{} achieves state-of-the-art performance without any high-resolution training data.

\section*{ACKNOWLEDGMENT}

This work is supported by Shimizu Corporation. Special thanks to Hayashi Daisuke regarding on-site experiments.


\bibliographystyle{IEEETran}
\bibliography{ref}
\end{document}